\documentclass[letterpaper, 10 pt, conference]{ieeeconf}

\IEEEoverridecommandlockouts
\usepackage{cite}
\usepackage{float}
\usepackage{graphicx}
\usepackage{amssymb}

\usepackage{amsmath}
\usepackage{mathtools}  

\usepackage{nccmath}

\usepackage{amsthm}
\usepackage{tabularray}
\usepackage{booktabs}
\usepackage{makecell}

\usepackage[usenames,dvipsnames,svgnames,table]{xcolor}
\theoremstyle{plain}

\usepackage{float}
\floatstyle{plaintop}
\usepackage{MnSymbol}%
\usepackage{wasysym}%
\usepackage{caption}
\usepackage{subcaption}
\usepackage{graphicx}
\usepackage{import}
\usepackage[hidelinks]{hyperref}

\renewcommand{\baselinestretch}{0.96}

\usepackage{algorithm}
\usepackage{algpseudocode}

\definecolor{deepblue}{rgb}{0,0,1}
\usepackage{dblfloatfix}

\definecolor{SpringGreen}{RGB}{40,160,80} 
\definecolor{SkyBlue}{RGB}{60,130,200}   
\definecolor{Khaki}{RGB}{160,100,70}     
\definecolor{Lavender}{RGB}{180,100,180}  
\definecolor{FruitOrange}{RGB}{245,218,95}

\begin{document}

\title{ 
\bf 
MochiSwarm: A testbed for robotic blimps in realistic environments
}

\author{
Jiawei Xu, Thong Vu, Diego S. D'Antonio, and David Salda\~{n}a
\thanks{J. Xu T. Vu, D. S. D'Antonio, and D. Salda\~{n}a are with the Autonomous and Intelligent Robotics Laboratory (AIRLab), Lehigh University, PA, 18015, USA:\texttt{\{jix519, tdv225, diego.s.dantonio, saldana\}@lehigh.edu}}
\thanks{The authors acknowledge the contributors and developers of this open-source platform: Edward Jeffs, Subhadeep Koley, Leonardo Santens, Shuhang Hou, Tongshu Wu, Karen Li, Hanqing Qi.}
}

\maketitle

\begin{abstract}
Testing aerial robots in tasks such as pickup-and-delivery and surveillance significantly benefits from high energy efficiency and scalability of the deployed robotic system. 
This paper presents MochiSwarm, an open-source testbed of light-weight robotic blimps, ready for multi-robot operation without external localization. We introduce the system design in hardware, software, and perception, which capitalizes on modularity, low cost, and light weight. The hardware allows for rapid modification, which enables the integration of additional sensors to enhance autonomy for different scenarios. The software framework supports different actuation models and communication between the base station and multiple blimps. The detachable perception module allows independent blimps to perform tasks that involve detection and autonomous actuation. We showcase a differential-drive module as an example, of which the autonomy is enabled by visual servoing using the perception module. A case study of pickup-and-delivery tasks with up to $12$ blimps highlights the autonomy of the MochiSwarm without external infrastructures.
\end{abstract}

\section{Introduction}{
The field of aerial robotics is rapidly advancing, with significant progress in applications ranging from disaster response and environmental monitoring to entertainment~\cite{asadzadeh2022uav,alvissalim2012swarm,kim2018survey}. However, deploying aerial systems in realistic settings remains a key challenge due to environmental unpredictability~\cite{Patino2023}. Unlike other types of aerial vehicles that actively compensate for gravity, blimps leverage natural buoyancy from gases such as helium to elongate their flight duration~\cite{Petrescu2017} and counteract gravity~\cite{680971} without continuous power consumption. In addition, the typical flexibility of LTA gas containers makes them inherently safer in collisions, ideal for environments with humans or sensitive equipment~\cite{9517490}. 

On the downside, blimps suffer from limited agility due to their size. Forcing high-speed operation with powerful actuators often leads to wasted energy in air drag~\cite{batchelor2000introduction}. Rather than enhancing individual vehicle performance, deploying a multitude of blimps is an effective alternative. The  support for multi-robot operation not only bypasses the diminishing marginal returns of improving individual blimps, but also suits the system better in certain scenario such as multi-target pickup-delivery~\cite{9410352}, and environmental monitoring~\cite{680971,s22051824}, as the operation could distribute the task into parallel sub-routines for individual robots~\cite{choudhury2021efficient}.

\begin{figure}[t]
  \centering
  \includegraphics[trim={0cm 1.5cm 0.0cm 0.0cm},clip,width=0.9\linewidth]{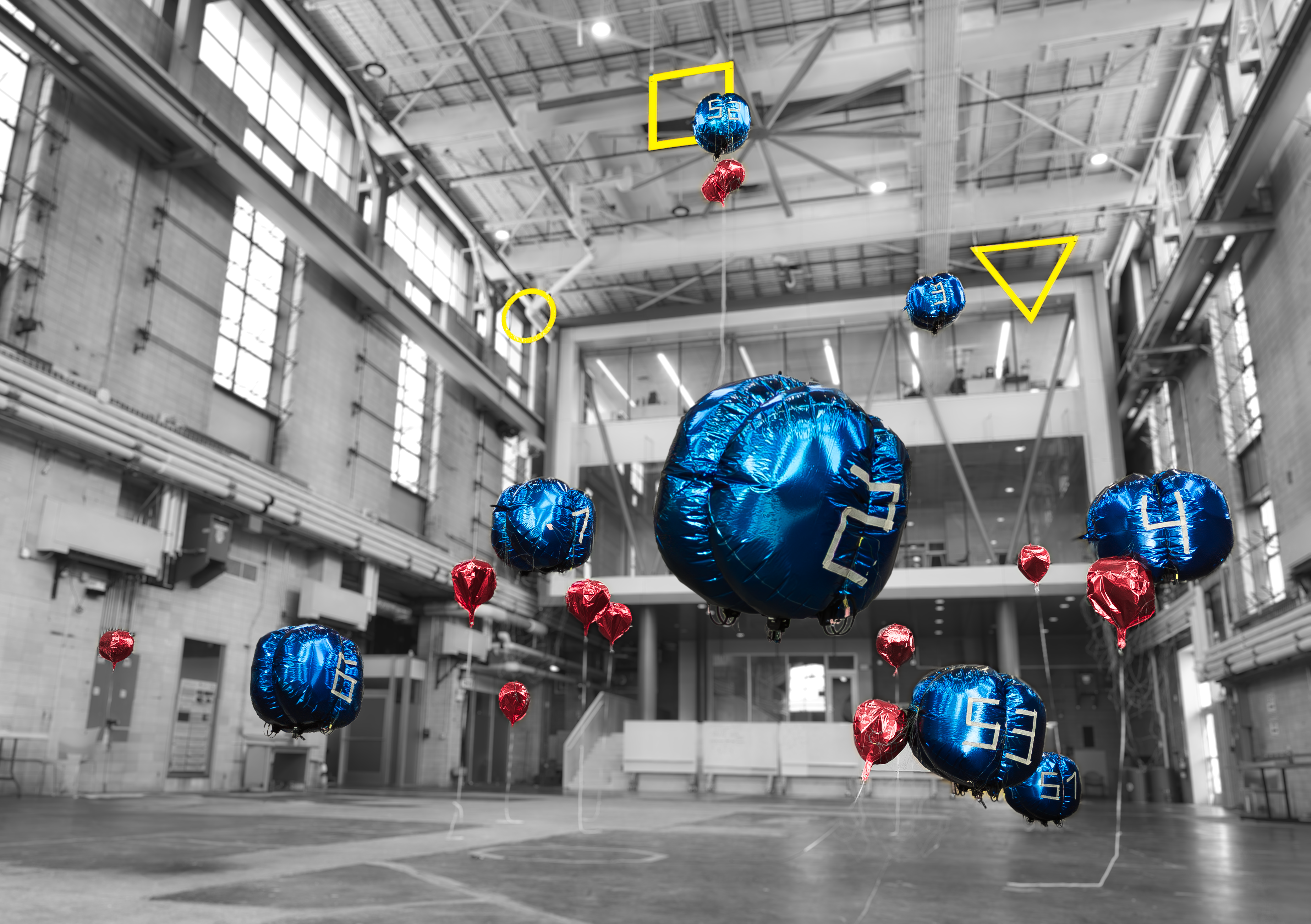}
  \caption{We test the MochiSwarm system in an industrial highbay. The blue blimps capture red target balloons and deliver them through light-yellow hoops. A video summary is available at \url{https://youtu.be/75gfIlPtrBU}.}
  \vspace{-1em}
  \label{fig:team}
\end{figure}
Recent literature exhibits robotic blimp designs for practical applications~\cite{pellegrino2024tinyblimp,lin2022miniature}, some considering adversarial and multi-robot coordination~\cite{meighan2025des}, while a scalable system emphasizing versatility and modularity is yet to be introduced to accommodate diverse scenarios. We present MochiSwarm system to fill the gap, which is an open-source, modular, and low-cost robotic blimp platform tailored for swarm operations. Unlike existing multi-robot aerial systems such as Crazyswarm~\cite{7989376} and GRASP micro-UAV testbed~\cite{5569026} that rely on motion capture systems for precise localization, MochiSwarm is intended to operate autonomously without external infrastructures. The autonomy allows each blimp to take actions based on onboard sensors and the perception module. The communication design supports multi-robot operations without inducing significant overhead, addressing scalability challenges in centralized multi-robot systems~\cite{gul2022centralized}. In comparison to similar aerial swarm platforms that do not rely on external localization such as the swarm-VIO~\cite{8276634}, although individual blimps of MochiSwarm are vulnerable to environmental hazards, their low cost and long flight time offset the negative impact of possible failures.
MochiSwarm offers a novel multi-robot testbed of testing aerial vehicles in realistic scenarios that involve distributed computation, wind disturbance compensation, and indoor environments without an external localization system. The testbed is fully documented and open-source, and all components are commercially available, making the platform accessible for all in terms of construction and deployment. The modular design allows rapid hardware and software modifications for different tasks, advocating for community collaboration. Table.~\ref{tab:comparison} summarizes the features of MochiSwarm compared to established multi-robot aerial platforms.
\begin{table*}[b]
    \centering
    \caption{The comparison between MochiSwarm, Crazyswarm, GRASP micro-UAV testbed, and VIO-Swarm.}
    \begin{tabular}{|c|c|c|c|c|}
    \hline
                & \textbf{MochiSwarm}               & Crazyswarm~\cite{7989376}                             & GRASP testbed~\cite{5569026}      & VIO-Swarm~\cite{8276634}\\
    \hline
    Agent type  & Robotic blimp                     & Crazyflie Quadrotor                                   & Hummingbird quadrotor             & Qualcomm Dragonfly   \\
    \hline
    Open source & $\checkmark$                      & $\checkmark$                                          & $\times$                          & $\checkmark$\\
    \hline
    Autonomy    & Onboard Target Detection          & \makecell{External Localization\\ and Coordination}   & External Localization             & \makecell{Onboard Odometry\\ and External Coordination}\\
    \hline
    Extension   & I2C, development required         & Crazyflie Decks                                       & N/A                               & N/A \\
    \hline
    Scenario    & Indoor \& outdoor                 & Indoor                                                & Indoor                            & Indoor \& outdoor \\
    \hline
    Unit Cost   & \$132.74 + \$ 115 (NiclaVision)   & \$240 (w/o Motion Capture)                            & N/A                               & \$981.99 (w/o Motion Capture)\\
    \hline
    \end{tabular}
    \label{tab:comparison}
\end{table*}




\paragraph*{\textbf{Case Study}}
We design a multi-agent pickup-and-delivery (MAPD) task to evaluate MochiSwarm in a realistic environment. The mission takes place in an industrial highbay as shown in Fig.~\ref{fig:team}. The $20$ m $\times\ 15$ m area is filled with unorganized structures and a colorful background. The air conditioning units create an unknown wind turbulence up to $1$ m/s that affects the motion of the blimps. All factors emulate a challenging real-world situation. 
In the case study, the red balloons represent target objects, and three yellow retroreflective hoops close to the ceiling represent the goals that are circular, rectangular, and triangular. The objective is to locate and capture the target objects and then deliver them to a designated goal autonomously. Each blimp is equipped with a hanging net to capture targets. Lacking an external localization system for evaluation, we measure the performance of the system with the success rate in the task. This case study aims to test the effectiveness of swarm robots in covering large areas and adapting to uncertain conditions with limited sensing capabilities and computational power.

}

\section{Mochi Design}
\begin{figure}[t]
  \centering
  \includegraphics[width=0.9\linewidth]{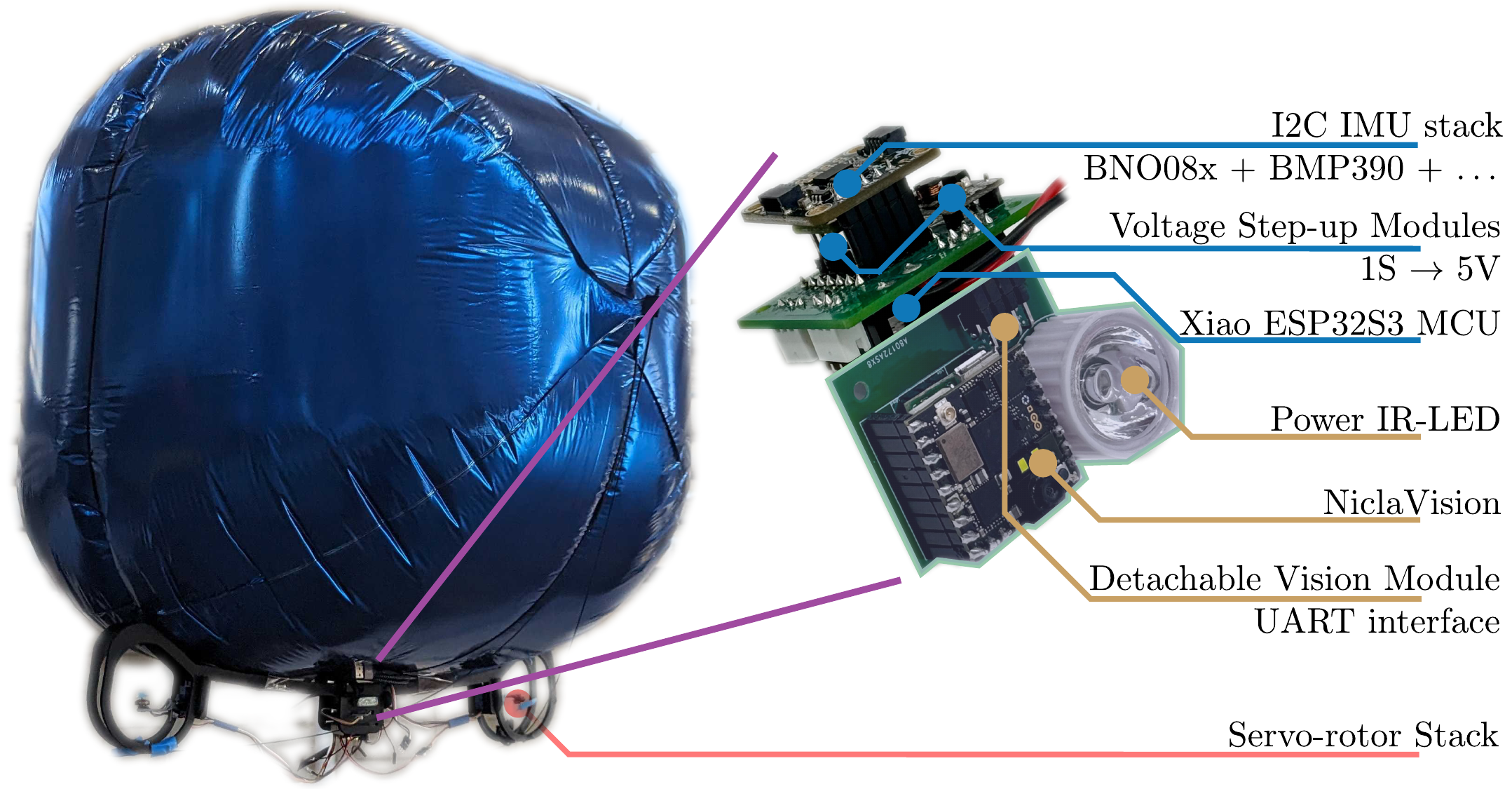}
  \caption{The hardware components of a Mochi blimp.}
  \vspace{-1em}
  \label{fig:mochi-vi}
\end{figure}
We design the MochiSwarm system so that it supports multi-robot coordination out-of-the-box. To facilitate future development and community collaboration, we emphasize three design principles, which are \emph{modularity}, \emph{low cost}, and \emph{light weight}. The modularity in both software and hardware allows to conveniently change components such as actuators and sensors to operate in specific scenarios. Combined with low-cost and open-source components, it enables rapid design iterations. By minimizing the weight of the structure and the onboard device, we allow the blimp to maintain neutral buoyancy with additional payloads or components. 

The three major subsystems of the MochiSwarm embody these principles. The \emph{hardware} which consists of electronic devices and a physical structure, the \emph{software} which implements the sensing, computation, control, and communication. The detachable \emph{perception} subsystem enables the blimps to detect target objects for pickup and delivery, tailored for the showcased MAPD task. 

\subsection{Electronics}
The electronics subsystem highlights components that are compact, widely accessible, and with moderate power, so that each blimp is light-weight and low-cost, while remaining compatible with additional equipment for different tasks. 

The circuit seats a XIAO ESP32S3 board as the center microcontroller unit (MCU) that integrates flight control, communication, and sensor fusion. The MCU features a compact form of $2.2$ cm $\times 1.7$ cm $\times 0.4$ cm, supports wireless communication, and is capable of running a flight controller at $200$ Hz on the 240-MHz dual-core processor.
Of the $11$ general-purpose input-outputs (GPIOs) available in the MCU, we reserve $4$ to generate pulse-width modulation (PWM) signals to drive actuators, $2$ for inter-integrated circuit (I2C) communication, $2$ for universal asynchronous receiver-transmitter (UART) communication, and $1$ for battery monitoring. On the custom printed circuit board (PCB), we connect the I2C pins to a set of header pins, facilitating the connection of additional sensors. We also create outlets for $2$ unused GPIOs for future development. 

For basic automation in the absence of the perception subsystem, we connect a barothermometer sensor and a 9-DoF magnetometer-accelerometer-gyroscope inertia measurement unit (IMU) to the PCB via the I2C bus. The barothermometer estimates the current above-sea altitude, which enables height control for the blimp. The IMU provides feedback on the heading direction with respect to the north and the tilting of the blimp in roll and pitch, allowing low-level explorative behaviors such as spiral search~\cite{fricke2016distributed}.

We showcase a differential-drive module for the Mochi blimp enabled by two actuator stacks, similar to a bicopter~\cite{li2023novel,he2022design}. Each stack is composed of a microservo and a 1-cell brushless rotor, located at the extremity of the frame to maximize torque generation while minimizing weight and inter-robot downwash interference, as shown in Fig.~\ref{fig:mochi-vi}. All actuators receive $50$-Hz PWM signals
to facilitate modifications to the actuation model, without the need to reconfigure the MCU output multiplexer.

The combination of sensors enable for a blimp with basic motion control. However, in MAPD tasks, the blimps need to detect the target objects and capture it. Visual servoing is a reliable approach to achieve high-level aerial autonomy~\cite{li2019modquad,lin2023open,taylor2025rem}. However, image processing induces a high computation overhead for microprocessors. Thus, to guarantee the performance of the flight control, we offload the visual servoing from the MCU to a detachable module based on NiclaVision. As a compact device, NiclaVision is capable of real-time image processing. Its compatibility with Arduino and Micropython aligns with our open-source design principle. We connect the NiclaVision to the main MCU using a custom PCB via UART protocol for bidirectional serial communication, as shown in Fig.~\ref{fig:mochi-vi}. In addition, we assign one NiclaVision GPIO to drive a $700$-mW infrared LED for retroreflective detection in the case study.

\begin{figure}[t]
    \centering
    \includegraphics[width=0.6\linewidth]{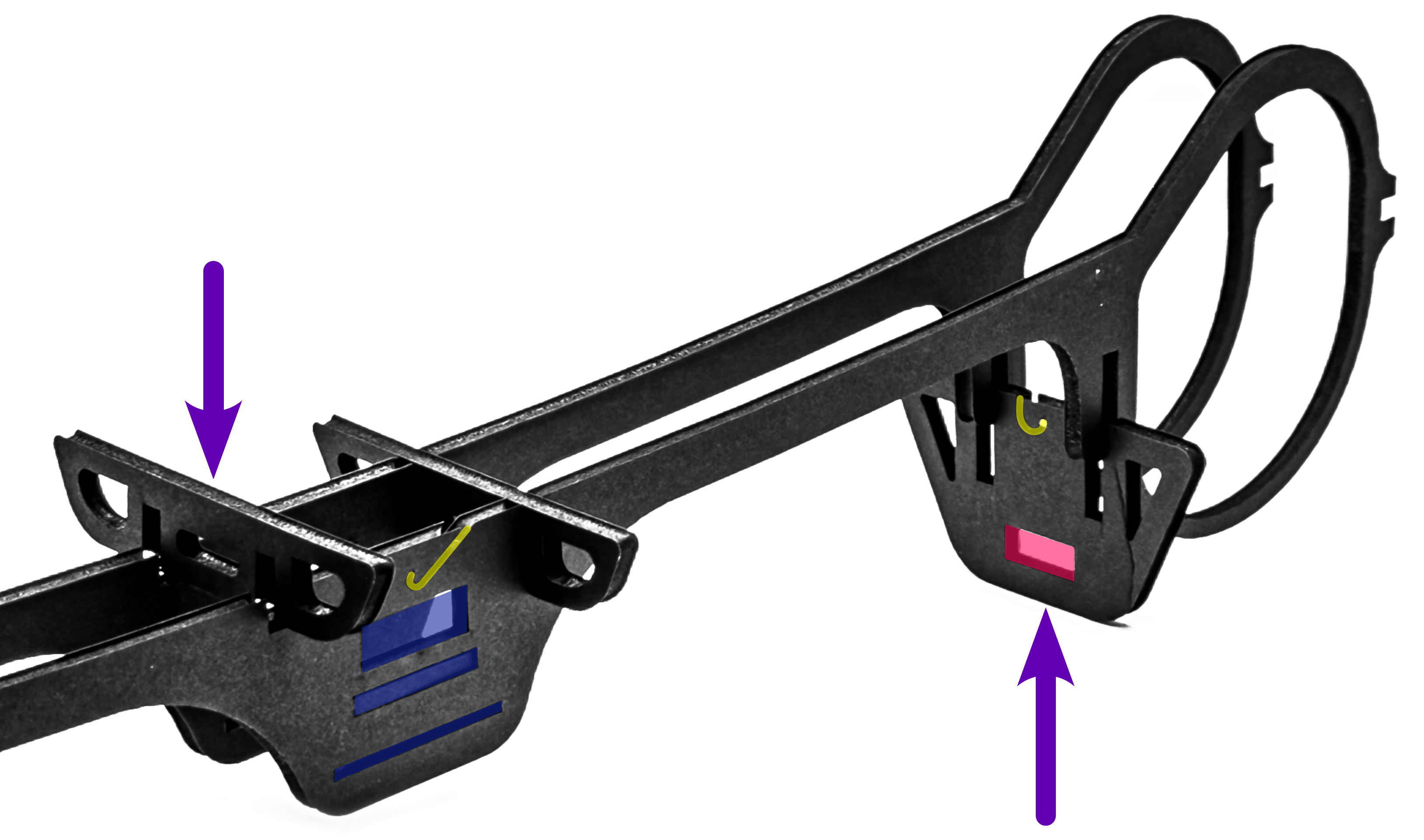}
    \caption{A close-up picture of the Mochi frame. The arrows show the direction of inserting the latch pieces to secure the frame, the central openings in blue host the battery, the sensor stack, and the main PCB from top to down, the red opening resides the servo-rotor stack, and the yellow traces host elastic strings that connect to the LTA balloon.}
    \label{fig:frame}
    \vspace{-1em}
\end{figure}
\begin{figure*}[t]
    \centering
    \includegraphics[width=0.95\linewidth,trim={0cm 0.25cm 0cm 0cm},clip]{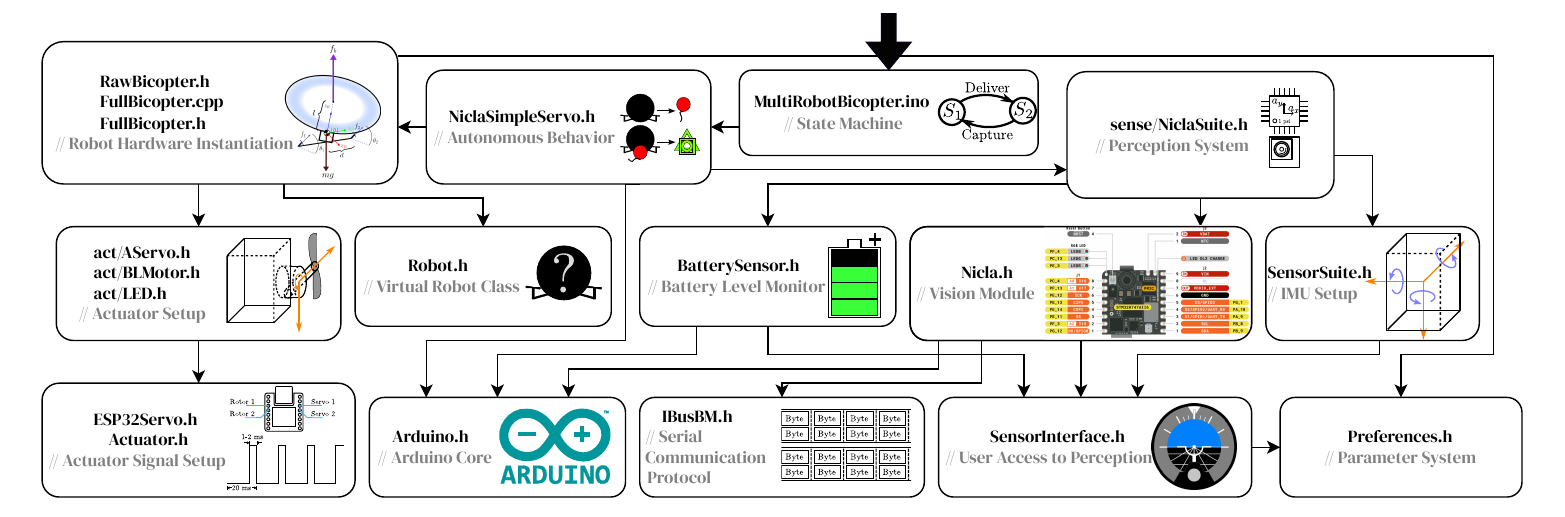}
    \vspace{-0.5em}
    \caption{The software architecture of the Mochi firmware. The bold arrow marks the entry of the firmware, which instantiates the state machine that specifies the autonomy of the blimp, depending on the task, the blimp actuation model, and the sensors available on the hardware construct. }
    \label{fig:software-architecture}
    \vspace{-1.5em}
\end{figure*}

\subsection{Structure}
Weight is a key factor to the payload capacity of a robotic blimp, which we optimize by increasing the buoyancy and decreasing the hardware weights.

The spherical foil balloon has $50$-inch diameter and contains helium gas that provides $120$ grams of net buoyancy. 
Its shape inspires the name ``Mochi'', a popular Japanese treat, and serves to evenly distribute lift and air drag during flight. 
The high balloon buoyancy enables the vehicle to hover with minimum thrust, making it ideal for extended operations.

In addition to choosing compact electronics and micro actuators, we reduce the weight of structural hardware by designing a minimalistic frame to hold together the components and the balloon. 
The frame is laser cut from a foam-core board, weighing $18.7$ g, which exhibits minimal deformation under external force below $2$~N. As shown in Fig.~\ref{fig:frame}, the frame assembly is achieved through interlock latch mechanisms with notches that match the thickness of the board. The frame locks together rigidly through friction without adhesives or screws, and attaches to the balloon via elastic strings. This design allows for rapid assembly and detachment of electronics, which simplifies component installation, improves modularity, and minimizes development downtime.

\subsection{Software}
For easier coordination of multiple blimps, we design the software architecture, including the onboard firmware and wireless communication, so that all blimps receive commands from a central device. To improve the scalability of this centralized architecture, we allow each blimp to run autonomously based on the state machine in the firmware \emph{without active communication} with the central device. 

The onboard firmware of the blimp handles flight control, sensor fusion, on-demand communication with the central device, and the state machine for autonomous behaviors. Its implementation in Arduino C enables rapid addition or removal of components\footnote{The fully documented software-hardware repository is available at~\url{https://github.com/LehighBlimpGroup/BlimpSwarm/}.}. 
The overview of the firmware architecture is shown in Fig.~\ref{fig:software-architecture}.
This choice over other software architectures such as FreeRTOS~\cite{arm2022measuring} and micro-ROS~\cite{belsare2023micro} helps us achieve the balance between the efforts required for development and firmware performance.


We modularize all sensor-related functions into independent libraries for the firmware. The firmware instantiates individual sensor libraries through the ``suite'' of sensors, implemented as a class. Each suite can contain any combination of sensors based on the hardware construction of a robot. Sensor readings are stored in a buffer array in the suite class accessible to other modules, \emph{e.g.,} the flight control.

For a specific task, we implement a state machine as an Arduino sketch file that includes states associated with singular behaviors, which is the entry point for the firmware that welcomes user modification.
In this work, our blimps run in four states. 
1) \texttt{Manual}: a human-operated state, \emph{accessible only via central command.}
2) \texttt{Random Walk}: a timed forward motion in a random direction. 
3) \texttt{MoveToGoal}, a visual-servoing state driving the blimp towards a target. 
4) \texttt{PassThroughGoal}, a visual-servoing state following \texttt{MoveToGoal} and driving the blimp to charge towards a target.  
We represent each state with a file that contains a behavior function and transitioning conditions to other states. In every control iteration, the firmware pulls the desired motion command from the behavior function of the current state, takes the sensor readings for feedback and state transitioning check, and computes the desired actuation. 

The MCU of a blimp communicates with onboard components through I2C and UART, and with other ESP32 MCUs through the ESP-NOW wireless protocol~\cite{pasic2021esp} that offers peer-to-peer communication up to $480$ meters and near-instantaneous data transmission at $512$ Kbps. 
We implement the transmitter on an ESP32 MCU connected to the central device to utilize these features. 
The data stream from the central device to the blimps sets parameters to be stored in the flash of blimp MCUs, which allows the user to tune the system behavior without re-flashing its firmware. The on-demand data stream from a blimp to the central device provides readings of individual blimps for user inspection.

\section{Perception}
The perception subsystem is essential for the MAPD task, which provides the detection of target balloons and goal hoops based on color and shape using a NiclaVision for high-level autonomy based on visual-servoing\footnote{The Micropython program using the OpenMV firmware for image capturing and processing is accessible at~\url{https://github.com/LehighBlimpGroup/Blob-detection-and-Tracking}.}.

\subsection{Balloon Detection}{
    In the case study, since the target balloons are of a specific color, a straightforward solution is to use color detection to locate them. However, irregular reflections caused by the aluminum balloon surface result in uneven color spread. When using a blob detector~\cite{4767053} to distinguish the balloons from the background, the bright areas can create separate subregions within a single balloon, leading to false positives. Object detectors such as Yolo~\cite{tang2024hic} can detect the balloon despite the colors. However, they are computationally heavy and cause high latency on a microprocessor. Therefore, we implement a color-family detector.
    \begin{figure}[t]
        \centering
        \includegraphics[height=0.35\linewidth]{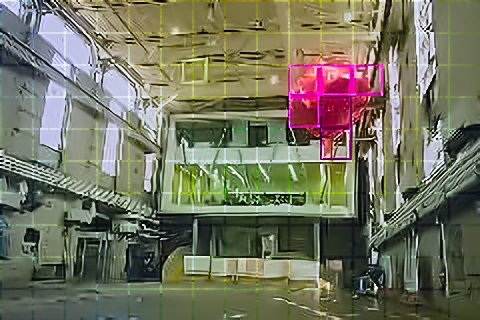}\hspace{1mm}
        \includegraphics[trim={1cm 0.5cm 2cm 0cm},clip,height=0.35\linewidth]{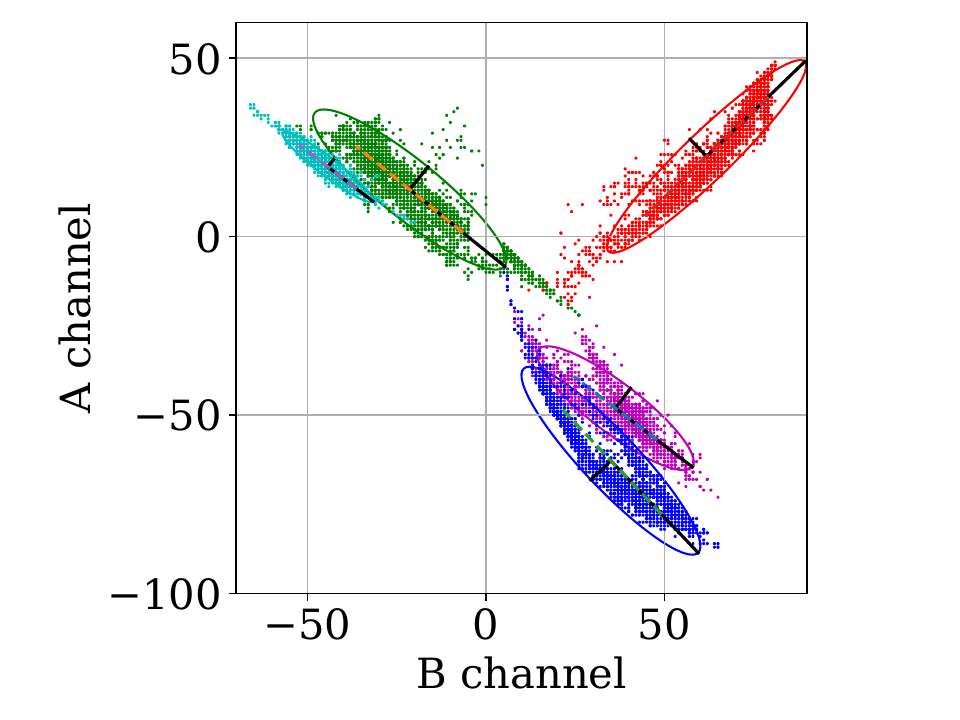}
        \caption{Left: the image frame captured by the NiclaVision where The pink blocks highlight the activated grid cells that match the balloon color. Right: the color distribution in the LAB color space for balloons with different colors.}
        \label{fig:color-balloon}
        \vspace{-1em}
    \end{figure}
    Our light-weight data-driven color detector\footnote{The code base for the color family generation is available at~\url{https://github.com/LehighBlimpGroup/balltraining}.} is based on segmented cells of the image frame.
    Before deployment, we collect images of target balloons in the environment using the NiclaVision. By segmenting each image into rectangular cells, we select the cells occupied by the balloons. Each of them serves as a data point in terms of its mean in the LAB color space~\cite{lissner2011toward}.     
    By projecting the data points across a large number of images in the A- and B-channels, we obtain the color dataset of the target balloon regardless of luminance as a cluster of points, as shown in the right half of Fig.~\ref{fig:color-balloon}. 
    The $2$-D Guassian distribution $\mathcal{N}_k(\boldsymbol{\mu}_k, \boldsymbol{\Sigma}_k)$ represents a color family $k$, with mean $\boldsymbol{\mu}_k$, and covariance~$ \boldsymbol{\Sigma}_k$ in the A- and B-channels.


    After deployment, the NiclaVision segments each captured image into the same rectangular cells as in the data collection, and calculates the mean color of each cell in A- and B-channels, $\boldsymbol{\mu}_c$, which is compared to the color family $k$ using the Mahalanobis distance~\cite{mclachlan1999mahalanobis} to its distribution $\mathcal{N}_k$,
    \begin{equation}
        d_{c,k}\left(\boldsymbol{\mu}_c, \mathcal{N}_k\right) = \sqrt{\left(\boldsymbol{\mu}_c - \boldsymbol{\mu}_k\right)^\top\boldsymbol{\Sigma}_k\left(\boldsymbol{\mu}_c - \boldsymbol{\mu}_k\right)}.\label{eq:Mahalanob}
    \end{equation}
    If the distance is less than a threshold, we consider the cell ``activated'' by the target balloon, as shown in the left half of Fig.~\ref{fig:color-balloon}, where the pink squares depict the activation. 
    
    The low camera resolution and high noise causes noisy detection. Thus, on top of the detector, we run a Bayes filter, the log-odds-ratio filter~\cite{thrun2002probabilistic}. This choice over other types of filters, such as a Kalman filter~\cite{alatise2017pose} or an FIR filter~\cite{fricke2016distributed} reflects our emphasis on the numerical stability, for the logarithmic operation scales the probabilities into values of lower float precision, and memory usage, for the recursive operation eliminates the need of a buffer.
    Considering the cells in the image as a grid map, we combine the previous belief and detection to estimate the probability of each given color in the image. The center coordinates of the largest cluster of connected activated cells in the image frame $\boldsymbol{c}_b$ and the number of cells in the cluster $n_b$ are sent to the main MCU so to drive the blimp to approach the target.
    Our detection algorithm runs at $10$ Hz in NiclaVision. Other boards, such as OpenMV Cam H7, can run the same algorithm at $30$ Hz due to lower camera latency.
    
}
\subsection{Goal Detection}{
    After picking up an object, the blimps deliver it to designated goal areas for the MAPD task. We represent these areas with three hoops in triangular, circular, and rectangular shapes, respectively, that hang from the ceiling. These goals are wrapped with retroreflective tape in vivid colors to stand out from the background environment.
    In normal scenarios where the color of the goal hoop alone is distinguished from the background, we deploy classic color-based blob detection because the retroreflective tape guarantees that the surface color on the hoop is uniform. Our detection algorithm removes false positives in the background that are not circular, rectangular, or triangular, implemented by counting corners in the bounding box of a detected blob.

    In cases where the background is clustered with colors close to the goal or in low-light conditions where noises from the high camera gain drastically degrade the image quality, NiclaVision synchronizes image capture with a blinking power infrared LED on the perception board. 
    Without an infrared filter, the camera sensor is able to pick up the excessive infrared reflection from the retroreflective tape, even with low sensor gains. In a quick succession, two images $F_1$ and $F_2$ are captured with the LED on and off, respectively, to reduce the frame difference caused by motion. The algorithm then calculates the pixel-wise image difference that gives a new frame, highlighting only reflective objects in the image $F = \Vert F_1 - F_2\Vert$. By keeping pixels in $F$ that have a luminance larger than a threshold, we obtain the binary mask for potential retro-reflect goals in the image frame. The largest connected white-pixel blobs that pass the shape and size filter are considered the closest goal hoop. Lastly, the NiclaVision sends the center $\boldsymbol{c}_g$ and size $n_g$ of the bounding box in the image frame to the MCU. 
}

\section{Dynamics and Control}
\begin{figure}[t]
    \centering
    \includegraphics[width=0.6\linewidth]{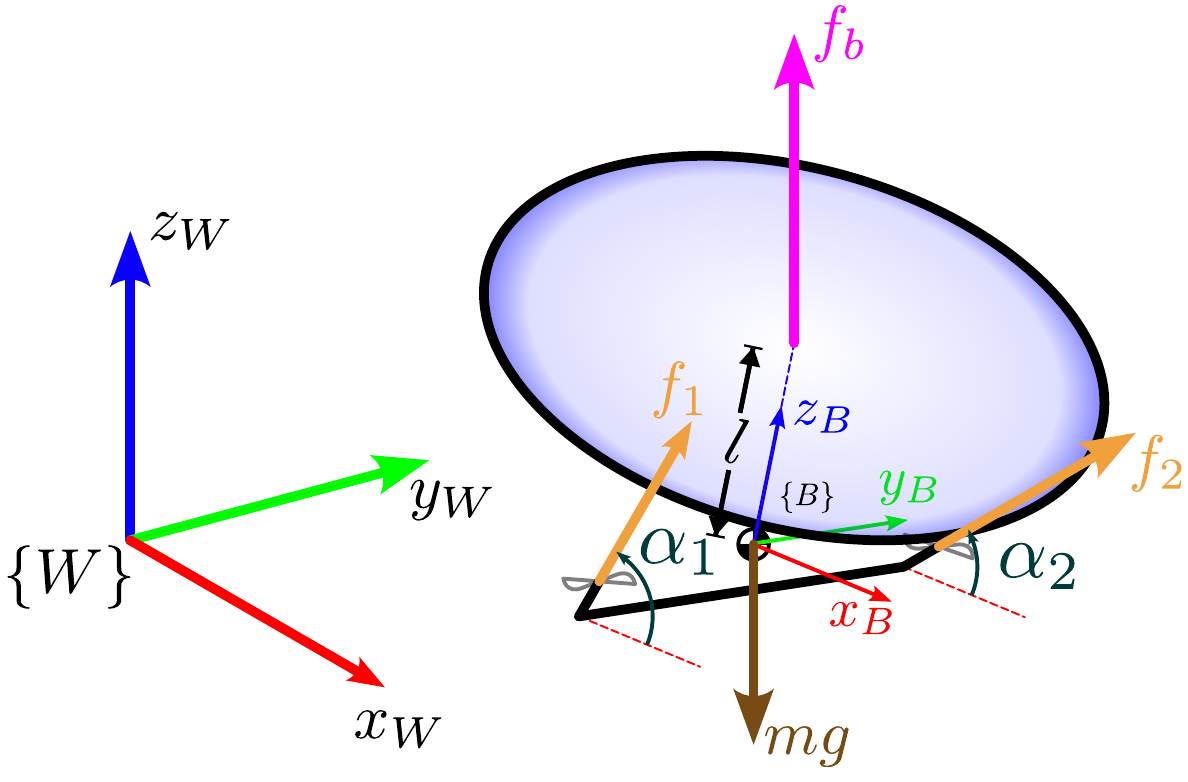}
    \caption{The abstracted model of a Mochi blimp, where the dowel points represents the center of mass.}
    \label{fig:blimpmodel}
    \vspace{-1em}
\end{figure}
We design a hybrid feedback-feedforward flight controller based on the IMU readings and the visual-servoing input for the fully autonomous operation of individual blimps. 

We define the world reference frame as a fixed frame, denoted by $\{W\}$ with its $x$-axis aligned with the magnet north and $z$-axis with the upward direction. The blimp has a body frame $\{B\}$ whose origin is at the center of mass (COM). Its $x$ axis points towards the front of the blimp, and its $z$ axis points upward, as shown in Fig.~\ref{fig:blimpmodel}. A pair of servo-rotor stacks actuate the blimp motion to rotate and translate. The servo-rotor stacks are mounted at both ends of a support arm placed beneath the balloon in the fashion of a bicopter~\cite{7748042}, both keeping a distance of $d$ from $z_B$. The support arm is in parallel with the $y$-axis of $\{B\}$ with a distance of $l_b$ below the blimp COM. Therefore, we denote their mounting positions in $\{B\}$ as $\boldsymbol{p}_1 = \left[0, -d, l_b\right]^\top$ and $\boldsymbol{p}_2 = \left[0, d, l_b\right]^\top$, respectively. Each servo-rotor stack generates a thrust force in a variable direction. The rotor forces are $f_1$ and $f_2$, respectively, and the rotation angles of the servos in the direction of $y_B$ are $\alpha_1$ and $\alpha_2$, respectively. At rest, \textit{i.e.}, $\alpha_i = 0$, the force vector of the $i$-th rotor aligns with $x_B$.

The translation $\boldsymbol{r}\in\mathbb{R}^3$ and rotation $\boldsymbol{R}\in\mathsf{SO(3)}$ from $\{W\}$ to $\{B\}$ describe the position and orientation of the blimp, respectively. $\boldsymbol{R}$ can be obtained from the Euler angles
$\phi, \theta, \text{ and } \psi$ in roll, pitch, and yaw, respectively~\cite{1098430}.
According to~\cite{lissaman1983low}, at low speed, the low Reynold number~\cite{batchelor1967introduction} results in an air drag approximately proportional to the velocity. Thus, we incorporate air drag as first-order damping terms for translation and rotation with constant diagonal matrices $\boldsymbol{D}_f$ and $\boldsymbol{D}_\tau$, respectively, that are both positive definite.
We use Newton-Euler equation to describe the dynamics of the blimp,
\begin{align}
    m\boldsymbol{\ddot{r}} &=\boldsymbol{Rf} + \boldsymbol{f}_e -\boldsymbol{D}_f\boldsymbol{\dot r},\label{eq:newton}\\
    \boldsymbol{J\dot{\omega}} + \boldsymbol{\omega}\times\boldsymbol{J\omega} &=\boldsymbol{\tau} + \boldsymbol{\tau}_e -\boldsymbol{D}_\tau\boldsymbol{\omega},\label{eq:euler}
\end{align}
where the net force and torque vectors generated by the thrusters in $\{B\}$ are 
\begin{eqnarray}
    \boldsymbol{f} &=& \sum_{i=1}^{2}f_i\left[\cos{\alpha_i}, 0, \sin{\alpha_i}\right]^\top,\label{eq:forceB}\\
    \boldsymbol{\tau} &=& \sum_{i=1}^{2}f_i\left(\boldsymbol{p}_i\times\left[\cos{\alpha_i}, 0, \sin{\alpha_i}\right]^\top\right).\label{eq:torqueB}
\end{eqnarray}

The external force from gravity and the buoyancy is $\boldsymbol{f}_e = \left[0, 0, f_b - mg\right]^\top$, where $g$ is the gravitational acceleration, $f_b$ is the buoyant force, and $m$ is the total mass of the blimp. The external torque from the buoyancy is $\boldsymbol{\tau}_e = \left(\boldsymbol{R}\left[0, 0, l\right]\right)\times\left[0, 0, f_b\right]^\top$ and $\boldsymbol{J}$ is the matrix of inertia moment.

Without precise localization, we rely on the onboard sensor readings to obtain state feedback, which consists of the heading direction with respect to the magnet north~$\psi$, the roll and pitch from the horizontal plane $\phi$, and $\theta$, respectively, and the height from the sea level, $h$, including their derivatives. The controller consists of two parts, one for height and yaw stabilization and the other for satisfying egocentric motion commands.
Since the construction of the blimp preserves natural stability in roll and pitch~\cite{10341796}, the feedback controller actively stabilizes its height and yaw around the desired values. 
Given a desired height $h^d$ and yaw $\psi^d$, we apply PD control to obtain the desired linear and angular acceleration in $z$-direction and in yaw,
\begin{eqnarray}
    \boldsymbol{\Ddot{r}}^d &=& \left(ke_h + k_d\dot e_h\right)[0, 0, 1]^\top,\label{eq:height-control}\\
    \boldsymbol{\dot{\omega}}^d &=& \left(k_R e_\psi + k_{R, d}\dot e_\psi\right)[0, 0, 1]^\top,\label{eq:yaw-control}
\end{eqnarray}
where $e_h = h^d - h$ and $\dot e_h$ are the desired height change and the corresponding velocity change from the current value, $e_\psi = \psi^d - \psi$ and $\dot e_\psi$ are the desired yaw change and yaw velocity change, and $k, k_d, k_R, k_{R, d}$ are gain coefficients. 

For the MAPD task, we obtain the desired height and yaw changes from the visual-servoing provided by the perception subsystem. Denoting the image center as $\boldsymbol{c}_f$, the desired change in height and yaw are computed from the pixel difference to the image center,
\begin{equation}
    \begin{bmatrix}
         e_\psi & e_h
    \end{bmatrix}^\top = \boldsymbol{K}\left(\boldsymbol{c}_f - \boldsymbol{c}_d\right), \label{eq:detection-controller}
\end{equation}
where $\boldsymbol{K}$ is the diagonal conversion matrix that is positive definite, which relates the height and yaw changes to the pixel difference in the $y$- and $y$-axes of the image, repectively. $\boldsymbol{c}_d := \boldsymbol{c}_b$ when the current state is balloon capturing, and $\boldsymbol{c}_d := \boldsymbol{c}_g$ if it is goal approaching. The feedback tries to drive the blimp such that aligns its front-facing perception subsystem with the object.
While the feedback control allows the blimp to levitate and change yaw direction, the perception subsystem commands the blimp to move forward, which introduces a feedforward acceleration in the direction of $x_B$, $\Ddot{r}^x$, we fuse the two sources of actuation into the desired force and torque in $\{B\}$,
\begin{eqnarray}
    \boldsymbol{f}^d &=& \boldsymbol{R}^\top\left(m\boldsymbol{\Ddot{r}}^d - \boldsymbol{f}_e\right) + \Ddot{r}^x[1, 0, 0]^\top\label{eq:force-input},\\
    \boldsymbol{\tau}^d &=& \boldsymbol{J\dot{\omega}}^d + \boldsymbol{\omega}\times\boldsymbol{J\omega} - \boldsymbol{\tau}_e\label{eq:torque-input}.
\end{eqnarray}
This feedforward term represents the ``charge'' behavior in autonomy, and is set to a nonzero constant if the detected object is close, determined by the number of activated cells $n_b$ exceeding a threshold for balloon detection or the blob size $n_g$ surpassing a threshold for goal detection.
Note that the blimp is incapable of generating force in $y_B$ direction from~\eqref{eq:forceB}. We denote the first and third elements of the desired force $\boldsymbol{f}^d$ as $f_x$ and $f_z$, respectively, and the third element in $\boldsymbol{\tau}^d$ as $\tau_z$. 
Solving for the actuator inputs $f_i$ and $\alpha_i$ with the desired $f_x, f_z$, and $\tau_z$ from \eqref{eq:forceB} and \eqref{eq:torqueB}, we obtain 
\begin{equation}
    f_i = \sqrt{f_{ix}^2 + f_{iz}^2} \text{ and } \alpha_i = \arctan\left(\frac{f_{iz}}{f_{ix}}\right)\label{eq:actuation}
\end{equation}
where $f_{ix} = \frac{1}{2}\left(f_x +(-1)^i \frac{\tau_z}{d}\right) \text{ and }
    f_{iz} = \frac{1}{2}\left(f_z +(-1)^{i+1} \frac{\tau_x}{d}\right)$ for $i = 1, 2$.
    

\section{Experiments} {
    We conduct two sets of experiments on the MochiSwarm in MAPD tasks as the case study. The first set of experiments extensively tests the detection and capturing capability of the MochiSwarm with an increasing number of blimps in the swarm against an increasing number of target balloons. We would like to demonstrate the increasing workload capacity as the number of blimps in the swarm increases. The second set of experiments evaluates the full autonomous loop of pickup and delivery with a large number of blimps operating simultaneously as a team\footnote{The full video recordings of the experiments are available at~\url{https://tinyurl.com/MochiSwarm}.}. 

    \subsection{Pickup Test}
    We focus on the detection and capture aspects to test perception, motion, and the centralized multi-robot coordination. The red balloons are deployed in a center $5$ m $\times$ $5$ m area, and the blimps are initialized in a center $10$ m $\times$ $10$ m area with random position and direction.
    We increment the number of blimps from $1$ to $4$. and the number of red balloons from $1$ to $8$ in the environment until the number of pickup attempts stops increasing. For the metric, we count the number of attempted and successful pickups to find the workload capacity of the swarm. Each red balloon is tied to a weight so that it floats around a fixed height. Whenever the perception subsystem triggers a nonzero feedforward acceleration $\Ddot{r}^x > 0$, we consider it a pickup attempt. Whenever the net of a blimp encloses a balloon and lifts the attached weight from the ground, we count it as a successful pickup. Afterwards, a human operator removes the captured balloon from the blimp that captures, and redeploys both in the environment, so that the blimp continues the autonomous search and pickup operation.

    The results are shown in Fig.~\ref{fig:pickup-test}. As the number of blimps increases, the maximum number of attempted pickups increases near-linearly as a result of increased workload capacity. Even with minimum coordination, the return of running a multitude of robots simultaneously is scalable. In situations where the target balloons are less than blimps, the ratio of successful pickups is sometimes smaller with a higher number of blimps. We trace the cause of this issue to the limited swarm coordination, since the unawareness of peer blimps may lead multiple blimps to collide with each other while chasing the same target. Third, the overall ratio of successful pickups remains the same for the swarms with different numbers of blimps when the targets are abundant, which depends on the individual detection and capturing capabilities. It contributes to an increasing number of successful pickups, almost linear to the number of blimps in the system, which once again shows that the performance of the swarm is scalable and not affected by a diminishing margin that is common when resources are dedicated to improve individual blimp competence.

    \begin{figure}
        \centering
        \includegraphics[width=0.45\linewidth]{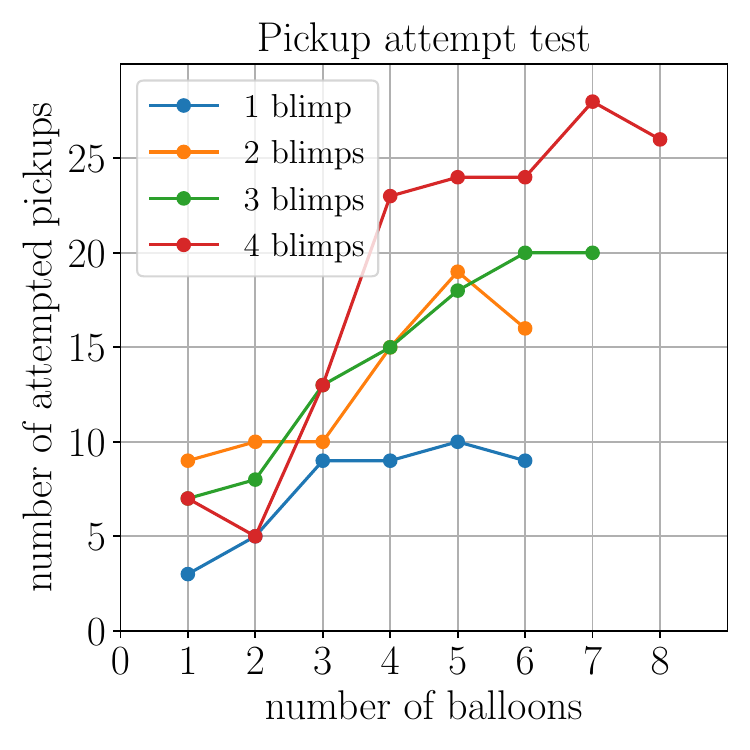}\hspace{1mm}\includegraphics[width=0.45\linewidth]{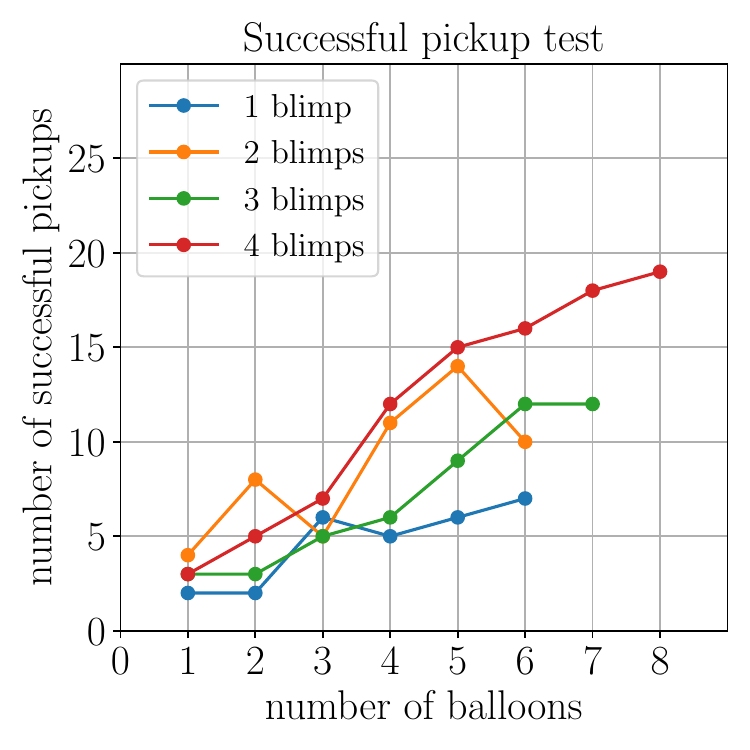}
        \caption{The number of attempted and successful pickups with increasing number of balloons and blimps in the swarm.}
        \label{fig:pickup-test}
        \vspace{-1em}
    \end{figure}
    \subsection{Pickup-and-delivery Test}
    We deploy a MochiSwarm of $4$ blimps to pick up and deliver $8$ and $12$ balloons in the first two full-loop tests, then a swarm of $8$ blimps for $12$ balloons in the last test. The blimps autonomously pick up the target balloons and approach the goal hoops with the captured balloons in the net. Whenever a blimp with a balloon inside its net successfully makes contact with or passes through a goal, we count a successful delivery and manually reset the blimp to the ground. Each of the tests lasts $5$ minutes, and serves as a proof of concept for MochiSwarm. $4$ blimps deliver successfully $2$ of the $8$ targets and $3$ of the $12$ targets, and the $8$ blimps deliver successfully $5$ of the $12$ targets. During tests, the manual communication switch to any of the $8$ blimps is instantaneous, showcasing the competence of our communication architecture. Although individual blimps can get stuck near the AC units, the cables that connect to the goal hoop, and uneven wall structures, the swarm continues to tackle the task. The independent autonomy of individual blimps grants a performance scalable with the number of blimps.
}

\section{Conclusion}{
In this paper, we introduced the design and prototype of MochiSwarm, a robotic blimp testbed that supports swarm operation for complex tasks in challenging environments. We capitalized its design principles of modularity, low cost, and light weight, providing details on the hardware and software implementation of individual blimps, as well as the communication architecture for multi-robot coordination. We engineered a case study emulating real-world MAPD missions to test the MochiSwarm. We also designed a detachable perception subsystem to provide detections for individual blimps to achieve high-level autonomy. In experiments, we demonstrated that each blimp in the MochiSwarm operates independently, leading to scalable swarm performance for the MAPD task. In the future, we would like to further improve the reliability of MochiSwarm by enhancing the swarm coordination by enabling inter-robot communication, allowing detection results to be shared between blimps to achieve collective behaviors.
}
    

\bibliographystyle{IEEEtran}
\bibliography{ref.bib}

\begin{thebibliography}{10}
\providecommand{\url}[1]{#1}
\csname url@samestyle\endcsname
\providecommand{\newblock}{\relax}
\providecommand{\bibinfo}[2]{#2}
\providecommand{\BIBentrySTDinterwordspacing}{\spaceskip=0pt\relax}
\providecommand{\BIBentryALTinterwordstretchfactor}{4}
\providecommand{\BIBentryALTinterwordspacing}{\spaceskip=\fontdimen2\font plus
\BIBentryALTinterwordstretchfactor\fontdimen3\font minus \fontdimen4\font\relax}
\providecommand{\BIBforeignlanguage}[2]{{%
\expandafter\ifx\csname l@#1\endcsname\relax
\typeout{** WARNING: IEEEtran.bst: No hyphenation pattern has been}%
\typeout{** loaded for the language `#1'. Using the pattern for}%
\typeout{** the default language instead.}%
\else
\language=\csname l@#1\endcsname
\fi
#2}}
\providecommand{\BIBdecl}{\relax}
\BIBdecl

\bibitem{asadzadeh2022uav}
S.~Asadzadeh, W.~J. de~Oliveira, and C.~R. de~Souza~Filho, ``Uav-based remote sensing for the petroleum industry and environmental monitoring: State-of-the-art and perspectives,'' \emph{Journal of Petroleum Science and Engineering}, vol. 208, p. 109633, 2022.

\bibitem{alvissalim2012swarm}
M.~S. Alvissalim, B.~Zaman, Z.~A. Hafizh, M.~A. Ma'sum, G.~Jati, W.~Jatmiko, and P.~Mursanto, ``Swarm quadrotor robots for telecommunication network coverage area expansion in disaster area,'' in \emph{2012 Proceedings of SICE Annual Conference (SICE)}.\hskip 1em plus 0.5em minus 0.4em\relax IEEE, 2012, pp. 2256--2261.

\bibitem{kim2018survey}
S.~J. Kim, Y.~Jeong, S.~Park, K.~Ryu, and G.~Oh, ``A survey of drone use for entertainment and avr (augmented and virtual reality),'' \emph{Augmented Reality and Virtual Reality: Empowering Human, Place and Business}, pp. 339--352, 2018.

\bibitem{Patino2023}
D.~Patiño, S.~Mayya, J.~Calderon, K.~Daniilidis, and D.~Saldaña, ``Learning to navigate in turbulent flows with aerial robot swarms: A cooperative deep reinforcement learning approach,'' \emph{IEEE Robotics and Automation Letters}, vol.~8, no.~7, pp. 4219--4226, 2023.

\bibitem{Petrescu2017}
\BIBentryALTinterwordspacing
R.~V.~V. Petrescu, R.~Aversa, B.~Akash, J.~Corchado, S.~P. Kozaitis, T.~M. Abu-Lebdeh, A.~Apicella, and F.~I.~T. Petrescu, ``Airlander,'' \emph{Journal of Aircraft and Spacecraft Technology}, vol.~1, pp. 119--148, Jul 2017. [Online]. Available: \url{https://thescipub.com/abstract/jastsp.2017.119.148}
\BIBentrySTDinterwordspacing

\bibitem{680971}
A.~Elfes, S.~Siqueira~Bueno, M.~Bergerman, and J.~Ramos, ``A semi-autonomous robotic airship for environmental monitoring missions,'' in \emph{Proceedings. 1998 IEEE International Conference on Robotics and Automation}, vol.~4, 1998, pp. 3449--3455 vol.4.

\bibitem{9517490}
Y.~H. Pheh, S.~K.~H. Win, and S.~Foong, ``Spiced: Design and control of a safe spherical blimp using coandă effect,'' in \emph{2021 IEEE/ASME International Conference on Advanced Intelligent Mechatronics (AIM)}, 2021, pp. 270--277.

\bibitem{batchelor2000introduction}
G.~K. Batchelor, \emph{An introduction to fluid dynamics}.\hskip 1em plus 0.5em minus 0.4em\relax Cambridge university press, 2000.

\bibitem{9410352}
Z.~Chen, J.~Alonso-Mora, X.~Bai, D.~D. Harabor, and P.~J. Stuckey, ``Integrated task assignment and path planning for capacitated multi-agent pickup and delivery,'' \emph{IEEE Robotics and Automation Letters}, vol.~6, no.~3, pp. 5816--5823, 2021.

\bibitem{s22051824}
\BIBentryALTinterwordspacing
A.~Fascista, ``Toward integrated large-scale environmental monitoring using wsn/uav/crowdsensing: A review of applications, signal processing, and future perspectives,'' \emph{Sensors}, vol.~22, no.~5, 2022. [Online]. Available: \url{https://www.mdpi.com/1424-8220/22/5/1824}
\BIBentrySTDinterwordspacing

\bibitem{choudhury2021efficient}
S.~Choudhury, K.~Solovey, M.~J. Kochenderfer, and M.~Pavone, ``Efficient large-scale multi-drone delivery using transit networks,'' \emph{Journal of Artificial Intelligence Research}, vol.~70, pp. 757--788, 2021.

\bibitem{pellegrino2024tinyblimp}
M.~Pellegrino, M.~Barci{\'s}, J.~Simonjan, Zulkarnaen, C.~F. Chiasserini, and E.~Natalizio, ``Tinyblimp: A promising frontier for autonomous miniature unmanned aerial vehicles,'' in \emph{Proceedings of the 10th Workshop on Micro Aerial Vehicle Networks, Systems, and Applications}, 2024, pp. 1--6.

\bibitem{lin2022miniature}
T.~X. Lin, M.~Rossouw, A.~B. Maxseiner, T.~Schuler, M.~A. Garratt, S.~Ravi, F.~Zhang, D.~M. Lofaro, and D.~A. Sofge, ``Miniature autonomous blimps for indoor applications,'' in \emph{AIAA SCITECH 2022 Forum}, 2022, p. 1834.

\bibitem{meighan2025des}
\BIBentryALTinterwordspacing
A.~Meighan, C.~Taylor, J.~Ponniah, and O.~D. Dantsker, \emph{Design of a Multi-Agent Simulation With Heterogeneous Agents, Adversaries, and Targets}. [Online]. Available: \url{https://arc.aiaa.org/doi/abs/10.2514/6.2025-2478}
\BIBentrySTDinterwordspacing

\bibitem{7989376}
J.~A. Preiss, W.~Honig, G.~S. Sukhatme, and N.~Ayanian, ``Crazyswarm: A large nano-quadcopter swarm,'' in \emph{2017 IEEE International Conference on Robotics and Automation (ICRA)}, 2017, pp. 3299--3304.

\bibitem{5569026}
N.~Michael, D.~Mellinger, Q.~Lindsey, and V.~Kumar, ``The grasp multiple micro-uav testbed,'' \emph{IEEE Robotics \& Automation Magazine}, vol.~17, no.~3, pp. 56--65, 2010.

\bibitem{gul2022centralized}
F.~Gul, A.~Mir, I.~Mir, S.~Mir, T.~U. Islaam, L.~Abualigah, and A.~Forestiero, ``A centralized strategy for multi-agent exploration,'' \emph{IEEE Access}, vol.~10, pp. 126\,871--126\,884, 2022.

\bibitem{8276634}
A.~Weinstein, A.~Cho, G.~Loianno, and V.~Kumar, ``Visual inertial odometry swarm: An autonomous swarm of vision-based quadrotors,'' \emph{IEEE Robotics and Automation Letters}, vol.~3, no.~3, pp. 1801--1807, 2018.

\bibitem{fricke2016distributed}
G.~M. Fricke, J.~P. Hecker, A.~D. Griego, L.~T. Tran, and M.~E. Moses, ``A distributed deterministic spiral search algorithm for swarms,'' in \emph{2016 IEEE/RSJ International Conference on Intelligent Robots and Systems (IROS)}.\hskip 1em plus 0.5em minus 0.4em\relax IEEE, 2016, pp. 4430--4436.

\bibitem{li2023novel}
K.~Li, S.~Hou, M.~Negash, J.~Xu, E.~Jeffs, D.~S. D'Antonio, and D.~Salda{\~n}a, ``A novel low-cost, recyclable, easy-to-build robot blimp for transporting supplies in hard-to-reach locations,'' in \emph{2023 IEEE Global Humanitarian Technology Conference (GHTC)}.\hskip 1em plus 0.5em minus 0.4em\relax IEEE, 2023, pp. 36--42.

\bibitem{he2022design}
X.~He and Y.~Wang, ``Design and trajectory tracking control of a new bi-copter uav,'' \emph{IEEE Robotics and Automation Letters}, vol.~7, no.~4, pp. 9191--9198, 2022.

\bibitem{li2019modquad}
G.~Li, B.~Gabrich, D.~Saldana, J.~Das, V.~Kumar, and M.~Yim, ``Modquad-vi: A vision-based self-assembling modular quadrotor,'' in \emph{2019 International Conference on Robotics and Automation (ICRA)}.\hskip 1em plus 0.5em minus 0.4em\relax IEEE, 2019, pp. 346--352.

\bibitem{lin2023open}
T.~X. Lin, T.~K. Schuler, D.~M. Lofaro, D.~Sofge, and F.~Zhang, ``The open-blimp: An open-source blimp platform for lighter-than-air research,'' in \emph{AIAA SCITECH 2023 Forum}, 2023, p. 0695.

\bibitem{taylor2025rem}
\BIBentryALTinterwordspacing
C.~Taylor, M.~Widjaja, and O.~D. Dantsker, \emph{Remotely-Processed Vision-Based Control of Autonomous Lighter-Than-Air UAVs With Real-Time Constraints}. [Online]. Available: \url{https://arc.aiaa.org/doi/abs/10.2514/6.2025-1344}
\BIBentrySTDinterwordspacing

\bibitem{arm2022measuring}
J.~Arm, O.~Ba{\v{s}}t{\'a}n, O.~Mih{\'a}lik, and Z.~Brad{\'a}{\v{c}}, ``Measuring the performance of freertos on esp32 multi-core,'' \emph{IFAC-PapersOnLine}, vol.~55, no.~4, pp. 292--297, 2022.

\bibitem{belsare2023micro}
K.~Belsare, A.~C. Rodriguez, P.~G. S{\'a}nchez, J.~Hierro, T.~Ko{\l}con, R.~Lange, I.~L{\"u}tkebohle, A.~Malki, J.~M. Losa, F.~Melendez \emph{et~al.}, ``Micro-ros,'' in \emph{Robot Operating System (ROS) The Complete Reference (Volume 7)}.\hskip 1em plus 0.5em minus 0.4em\relax Springer, 2023, pp. 3--55.

\bibitem{pasic2021esp}
R.~Pasic, I.~Kuzmanov, and K.~Atanasovski, ``Esp-now communication protocol with esp32,'' \emph{Journal of Universal Excellence}, vol.~6, no.~1, pp. 53--60, 2021.

\bibitem{4767053}
A.~J. Danker and A.~Rosenfeld, ``Blob detection by relaxation,'' \emph{IEEE Transactions on Pattern Analysis and Machine Intelligence}, vol. PAMI-3, no.~1, pp. 79--92, 1981.

\bibitem{tang2024hic}
S.~Tang, S.~Zhang, and Y.~Fang, ``Hic-yolov5: Improved yolov5 for small object detection,'' in \emph{2024 IEEE International Conference on Robotics and Automation (ICRA)}.\hskip 1em plus 0.5em minus 0.4em\relax IEEE, 2024, pp. 6614--6619.

\bibitem{lissner2011toward}
I.~Lissner and P.~Urban, ``Toward a unified color space for perception-based image processing,'' \emph{IEEE Transactions on Image Processing}, vol.~21, no.~3, pp. 1153--1168, 2011.

\bibitem{mclachlan1999mahalanobis}
G.~J. McLachlan, ``Mahalanobis distance,'' \emph{Resonance}, vol.~4, no.~6, pp. 20--26, 1999.

\bibitem{thrun2002probabilistic}
S.~Thrun, ``Probabilistic robotics,'' \emph{Communications of the ACM}, vol.~45, no.~3, pp. 52--57, 2002.

\bibitem{alatise2017pose}
M.~B. Alatise and G.~P. Hancke, ``Pose estimation of a mobile robot based on fusion of imu data and vision data using an extended kalman filter,'' \emph{Sensors}, vol.~17, no.~10, p. 2164, 2017.

\bibitem{7748042}
Q.~Zhang, Z.~Liu, J.~Zhao, and S.~Zhang, ``Modeling and attitude control of bi-copter,'' in \emph{2016 IEEE International Conference on Aircraft Utility Systems (AUS)}, 2016, pp. 172--176.

\bibitem{1098430}
R.~Pio, ``Euler angle transformations,'' \emph{IEEE Transactions on Automatic Control}, vol.~11, no.~4, pp. 707--715, 1966.

\bibitem{lissaman1983low}
P.~Lissaman, ``Low-reynolds-number airfoils,'' \emph{Annual review of fluid mechanics}, vol.~15, no.~1, pp. 223--239, 1983.

\bibitem{batchelor1967introduction}
G.~K. Batchelor, \emph{An introduction to fluid dynamics}.\hskip 1em plus 0.5em minus 0.4em\relax Cambridge university press, 1967.

\bibitem{10341796}
J.~Xu, D.~S. D'Antonio, D.~J. Ammirato, and D.~Saldaña, ``Sblimp: Design, model, and translational motion control for a swing-blimp,'' in \emph{2023 IEEE/RSJ International Conference on Intelligent Robots and Systems (IROS)}, 2023, pp. 6977--6982.

\end{thebibliography}
\end{document}